\documentclass[11pt]{article}
\usepackage[left=25mm, right=25mm, top=25mm, bottom=25mm, includehead=true, includefoot=true]{geometry}

\usepackage{graphicx}
\usepackage{url}
\usepackage{natbib} 
\usepackage{authblk} 
\usepackage[parfill]{parskip} 
\usepackage{enumitem} 

\usepackage{caption}
\usepackage{subcaption}
\usepackage[export]{adjustbox}  

\usepackage{sectsty}
\sectionfont{\normalsize}
\subsectionfont{\normalsize}
\subsubsectionfont{\normalsize}
\paragraphfont{\normalsize}

\usepackage[pdftex]{hyperref} 
\hypersetup{pdfborder={0 0 0} } 

\title{Probabilistic map-matching using particle filters}

\author[1]{Kira Kempinska\thanks{kira.kowalska.13@ucl.ac.uk}}
\author[1]{Toby Davies\thanks{toby.davies@ucl.ac.uk}}
\author[2]{John Shawe-Taylor\thanks{j.shawe-taylor@cs.ucl.ac.uk}}
\affil[1]{Department of Security and Crime Science, University College London}
\affil[2]{Department of Computer Science, University College London}

\begin{document}

\maketitle


\begin{abstract}
\centering

Increasing availability of vehicle GPS data has created potentially transformative opportunities for traffic management, route planning and other location-based services. Critical to the utility of the data is their accuracy. Map-matching is the process of improving the accuracy by aligning GPS data with the road network. In this paper, we propose a purely probabilistic approach to map-matching based on a sequential Monte Carlo algorithm known as particle filters. The approach performs map-matching by producing a range of candidate solutions, each with an associated probability score. We outline implementation details and thoroughly validate the technique on GPS data of varied quality.

$ $ 

\end{abstract}


\section{Introduction}

Over the last years we have witnessed a rapid increase in the availability of GPS-receiving devices, such as smart phones or car navigation systems. The devices generate vast amounts of temporal positioning data that have been proven invaluable in various applications, from traffic management \citep{Kuhne2003} and route planning \citep{Gonzalez2007,Li2011,Kowalska2015} to inferring personal movement signatures \citep{Liao2006}. 

Critical to the utility of GPS data is their accuracy. The data suffer from measurement errors caused by technical limitations of GPS receivers and sampling errors caused by their receiving rates. When digital maps are available, it is common practice to improve the accuracy of the data by aligning GPS points with the road network. The process is known as map-matching. 

Most map-matching algorithms align GPS trajectories with the road network by considering \textit{positions} of each GPS point, either in isolation or in relation to other GPS points in the same trajectory. The techniques, although computationally efficient, are not very accurate in cases when the sampling rate is low or the street network complexity is high. More advanced map-matching techniques utilise both \textit{timestamps} and \textit{positions} of GPS points in order to achieve a higher degree of accuracy. They would typically use temporal information to infer speed and then assign GPS points to roads that are in their proximity and which speed profiles best match the inferred speed. A prominent example of a spatio-temporal algorithm is ST-Matching \citep{Lou2009}. It has been shown to outperform purely spatial map-matching approaches, especially when the sampling rate is low.

The major limitation of both spatial and spatio-temporal approaches is their deterministic nature. They would always snap a GPS trajectory to a road network, regardless if it even came from the road network in the first place. The lack of confidence scores associated with their outputs might lead to very misleading results, especially when the data quality is low. 

In this paper, we address the issue of certainty by proposing a purely probabilistic spatio-temporal map-matching approach. It is based on a sequential Monte Carlo algorithm known as particle filters. The algorithm originates from the field of robotics \citep{Thrun2002}, where it has been widely applied in robot localisation problems. In the context of map-matching, it uses both spatial and temporal information to iteratively align a GPS trajectory with the road network; hence it can be used for both tracking and offline map-matching. It outputs the most likely road sequence that the GPS data came from together with the associated likelihood.

\section{Problem Statement}

In this section, we define the problem of probabilistic map-matching. 

Definition 1 (\textit{GPS trajectory}): A sequence of GPS points, where each GPS point contains latitude, longitude, bearing and timestamp. 

Definition 2 (\textit{Road network}): A directed graph with vertices representing road intersections and edges representing road segments. Bidirectional road segments are represented by two edges, each corresponding to a single direction of flow. Roads and intersections can be uniquely identified using their IDs.  

Definition 3 (\textit{Path}): A connected sequences of street segments in the road network.

\textit{Given a road network and a GPS trajectory, the goal of probabilistic map-matching is to find most probable paths that the GPS trajectory was generated from, together with their associated probability values. }

\begin{figure}[htbp] \begin{center} 
\resizebox{0.8\textwidth}{!}{ 
	\includegraphics{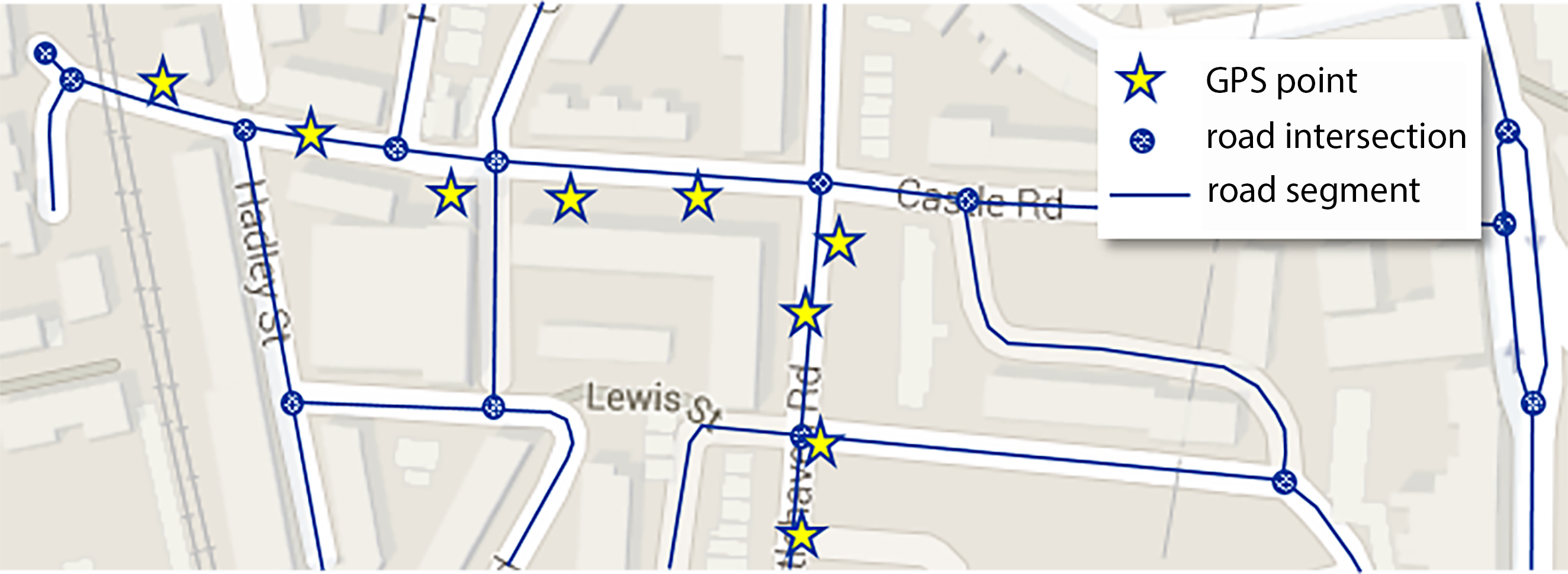}
} \caption{Exemplary road network with a GPS trajectory to be map-matched.} \label{first_figure} \end{center} \end{figure} %

\section{Methodology}
Our map-matching framework is based on particle filters. The algorithm computes candidate paths and their probabilistic values given a GPS trajectory. The most probable candidate path can then be selected as the map-matching outcome. The framework is evaluated using cross-validation.

\subsection{Particle Filter}
Particle filter is a sequential Monte Carlo technique that approximately infers true states of a dynamical system given its noisy observations. In our case, the dynamical system is a vehicle following a path along the road network, noisy observations are GPS points and the true states that we want to infer are actual locations of the vehicle at different timestamps. 

\begin{figure}[htbp] \begin{center} 
\resizebox{0.5\textwidth}{!}{ 
	\includegraphics{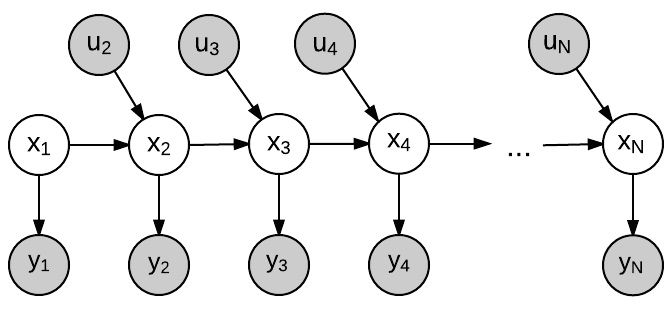}
} \caption{Graphical representation of a first-order Markov chain with hidden states $x_{1:N}$, measurements $y_{1:N}$ and controls $u_{1:N-1}$ at times $t=1:N$.} \label{hmm} \end{center} \end{figure}

The algorithm is based on the assumption that the dynamical system can be modeled as a first-order Markov chain with unobserved (hidden) states (see Figure~\ref{hmm}). That is, it assumes that the state of the system $x_t$ at time $t$ solely depends on the state at time $t-1$ through the so-called \textit{transition probability} $p(x_t|x_t-1,u_t)$, where $u_t$ is the control giving information about the change of the system in the time interval $(t-1;t]$. It adds that any measurements of the system are noisy descriptions of the unobserved true states, where the noise is modelled by the \textit{measurement probability} $p(y_t|x_t)$. The goal of particle filters is to infer $x_t$ given all available measurements $y_{1:t}$. The algorithm approximates the solution by recursively sampling from the posterior distribution \citep{Bishop2006}:

\begin{equation}
p(x_t \mid y^t,u^t )=const.\cdot p(y_t \mid x_t ) \int_  {} p(x_t \mid x_{t-1},u_t ) \cdot p(x_{t-1} \mid y^{t-1},u^{t-1} )
\end{equation}

under the initial condition $p(x_0 |y^0,u^0)=p(x_0)$ where $p(x_0)$ is the so-called \textit{initialisation distribution}. The samples are represented by \textit{particles}, i.e. possible states of the system given measurements. 

Definition 4 (\textit{Particle}): A point on the road network containing unique road segment identifier, distance along the segment and direction of travel (defined by from-to endpoints of the segment). 

The most basic version of particle filters is given by the following algorithm.

\begin{itemize}
\item \textbf{Initialisation}: At time $t=0$, draw M particles according to $p(x_0)$. Call this set of particles $X_0$.
\item \textbf{Recursion}: At time $t>0$, generate a particle $x_t$ for each particle in $X_{t-1}$ by sampling from the transition probability  $p(x_t \mid x_{t-1},u_t)$. Call the resulting set $\overline{X}_t$. Subsequently, draw $M$ particles (with replacement) with a probability proportional to the measurement probability $p(y_t \mid x_t)$. The resulting set of particles is $X_t$.
\end{itemize}

When the recursion reaches the last measurement at $t=N$, the particles stored in $X_N$ are approximate samples from the desired distribution $p(x_N \mid y_{1:N},u_{2:N})$. In our context, they represent possible paths taken by a vehicle given the GPS trajectory. The certainty associated with each path is proportional to the fraction of particles that it is represented by.

\subsection{Method Validation}

The easiest way to validate the accuracy of our map-matching approach would be to compare predicted paths with actual paths taken by a vehicle. Unfortunately, the ground truth is not available in our case study and we need validation techniques that overcome this limitation.

\begin{figure}[htbp] \begin{center} 
\resizebox{0.7\textwidth}{!}{ 
	\includegraphics{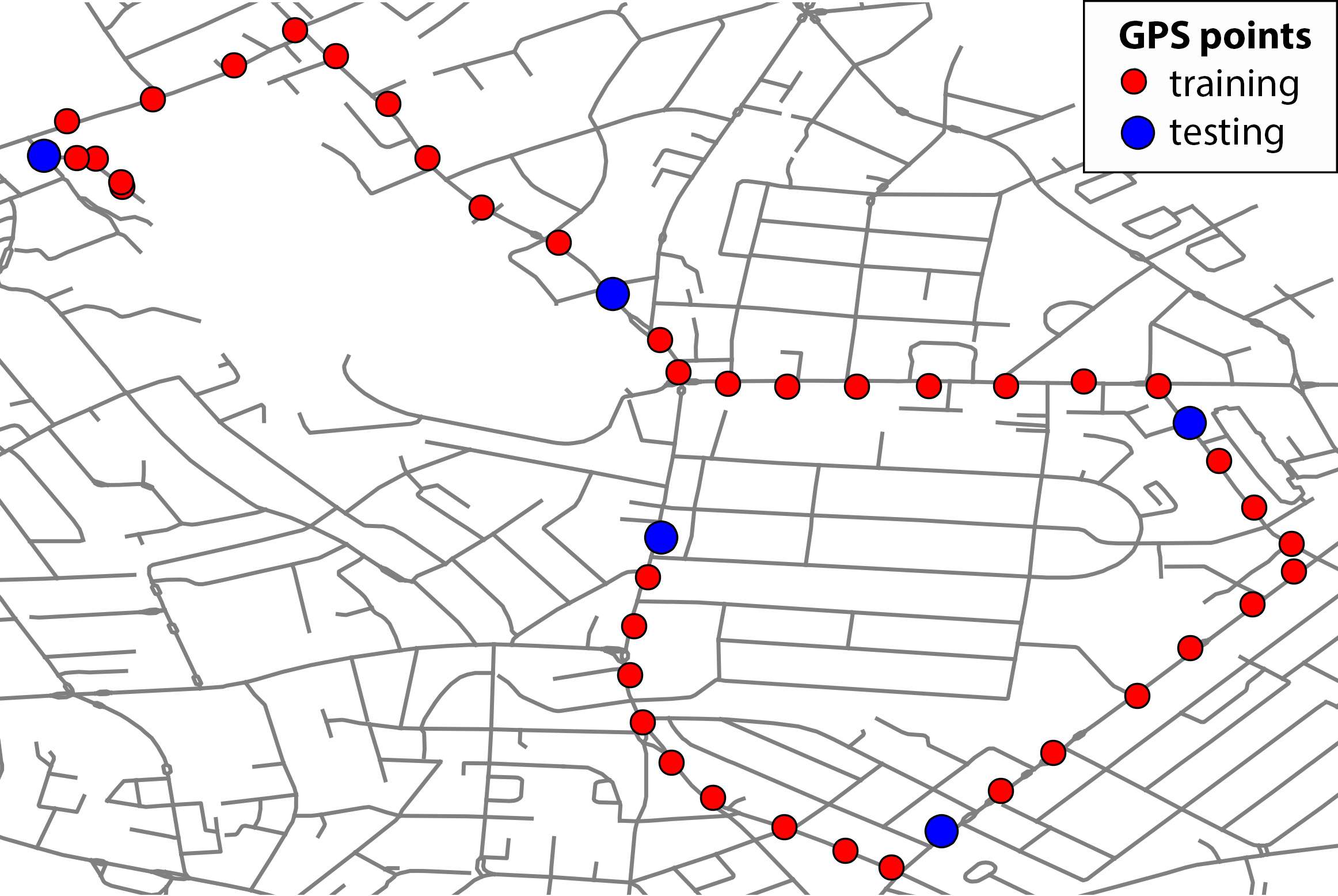}
} \caption{Exemplary GPS trajectory with points split into training and test sets.} \label{crossvalidation} \end{center} \end{figure}

We propose a validation framework based on the well-established technique of cross-validation \citep{Barber2012}. We remove 10\% of GPS points from each available GPS trajectory (see Figure~\ref{crossvalidation}). We then align the incomplete trajectories with the road network. Finally, we measure the distance between each removed point and the corresponding aligned path. The average distance across all removed points is our estimate of map-matching error.

\section{Results}

\subsection{Data}

The data motivating the project is a complete GPS trajectory of a police patrol vehicle during its night shift (9pm to 7am) in the London Borough of Camden on February 9\textsuperscript{th} 2015. The dataset contains 4,800 GPS points that were emitted roughly every second when moving. It was acquired for research purposes as part of the "Crime, Policing and Citizenship" project.\footnote{UCL Crime Policing and Citizenship: http://www.ucl.ac.uk/cpc/.}

\subsection{Implementation}

\begin{enumerate}[label=\Alph*]
\item \textbf{Initialisation} 

The initialisation probability distribution $p(x_0)$ is defined as a Gaussian centred at the position and bearing of the first GPS point. Particles initialised from the distribution are required to be positioned on the road network, hence their positions are first sampled (see Figure~\ref{pf_init_space}) and then either kept or discarded depending on whether they coincide with the road network or not (see Figure~\ref{pf_init_network}). Their direction of travel is inferred from the sampled bearing.

\begin{figure}
    \centering
    \begin{subfigure}[b]{0.45\textwidth}
        \includegraphics[width=\textwidth]{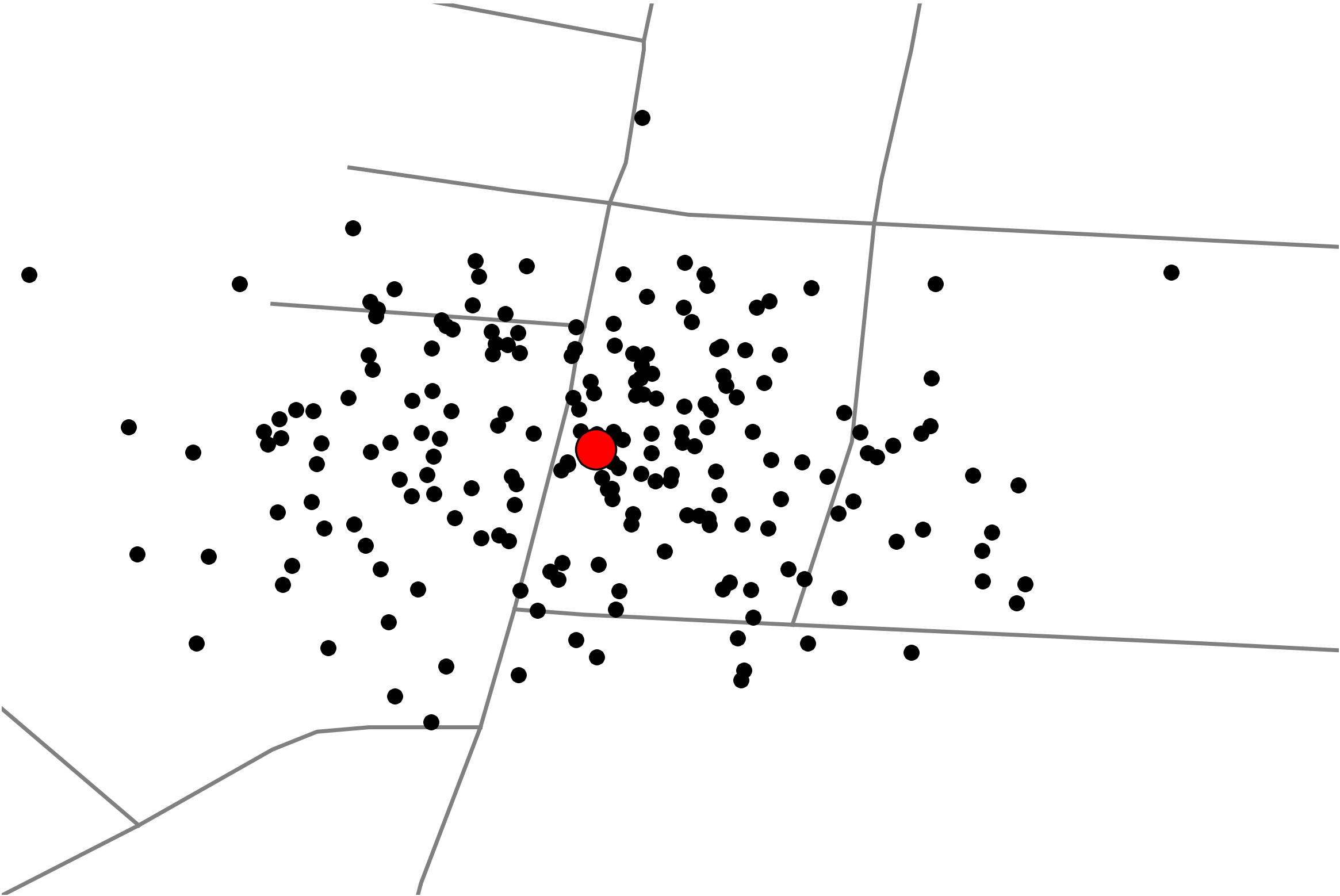}
        \caption{unconstrained}
        \label{pf_init_space}
    \end{subfigure}
    \hspace{2em}
    \begin{subfigure}[b]{0.45\textwidth}
        \includegraphics[width=\textwidth]{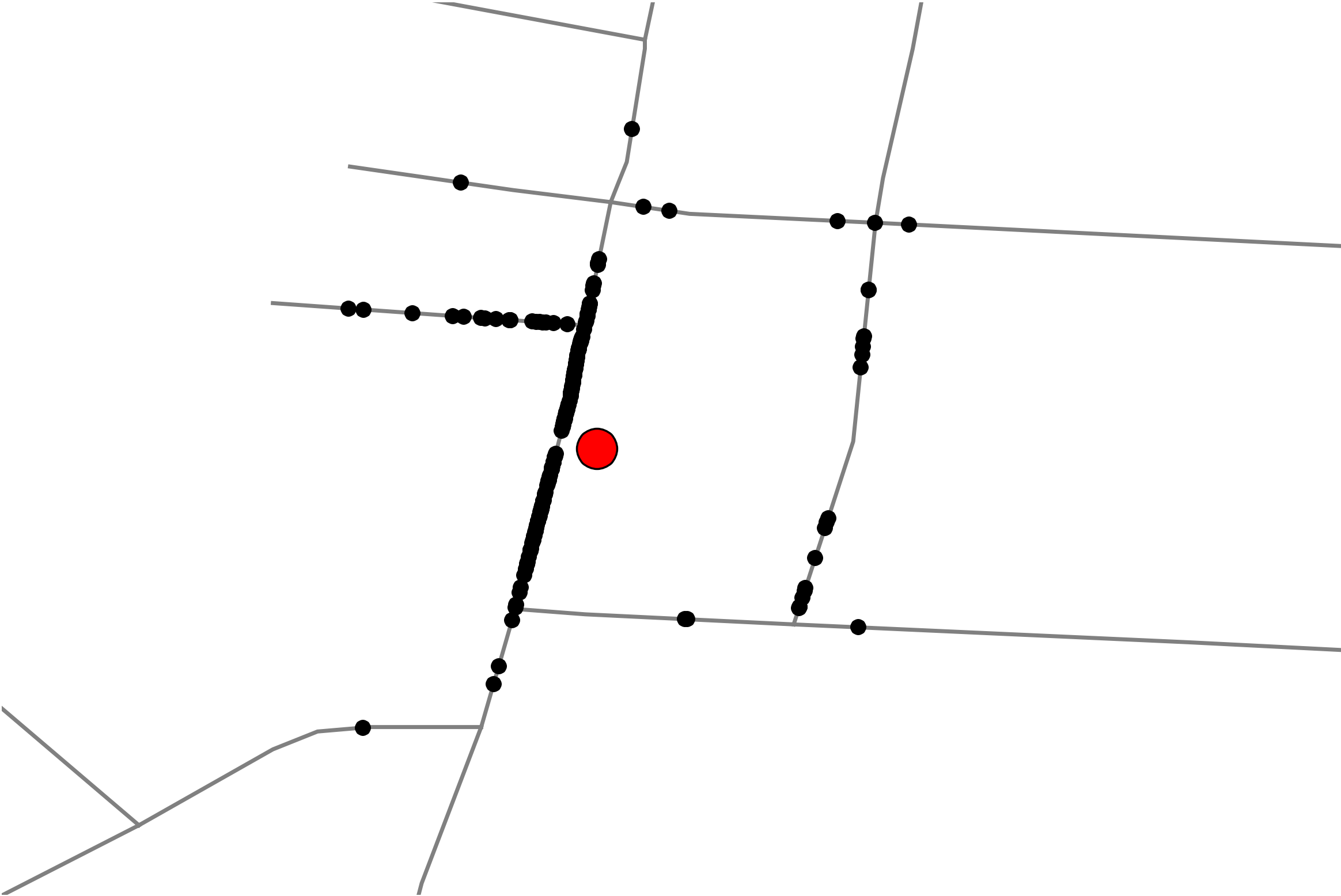}
        \caption{constrained to the road network}
        \label{pf_init_network}
    \end{subfigure}
    \caption{Initialisation of particles around the first GPS point in a trajectory.}\label{pf_init}
\end{figure}

\item \textbf{Transition probability} 

The transition probability $p(x_t \mid x_{t-1},u_t)$ is set as a linear estimate equal to the Cartesian distance between GPS points $x_{t-1}$ and $x_{t}$ (the control $u_t$) plus an additive Gaussian noise. 

In the recursive step of particle filter, particles move along the road network by a distance sampled from $p(x_ t \mid x_{t-1},u_t)$. When they encounter a road intersection, they randomly choose which road to follow.    

\item \textbf{Measurement probability}

Finally, the measurement noise $p(y_t \mid x_t)$ is also modelled as a Gaussian distribution, i.e. it is expected that GPS points are normally distributed around the true vehicle locations.
\end{enumerate}

\subsection{Performance Evaluation}

\begin{figure}
    \centering
    \begin{subfigure}[b]{0.45\textwidth}
    	\centering
        \includegraphics[height = 6 cm,right]{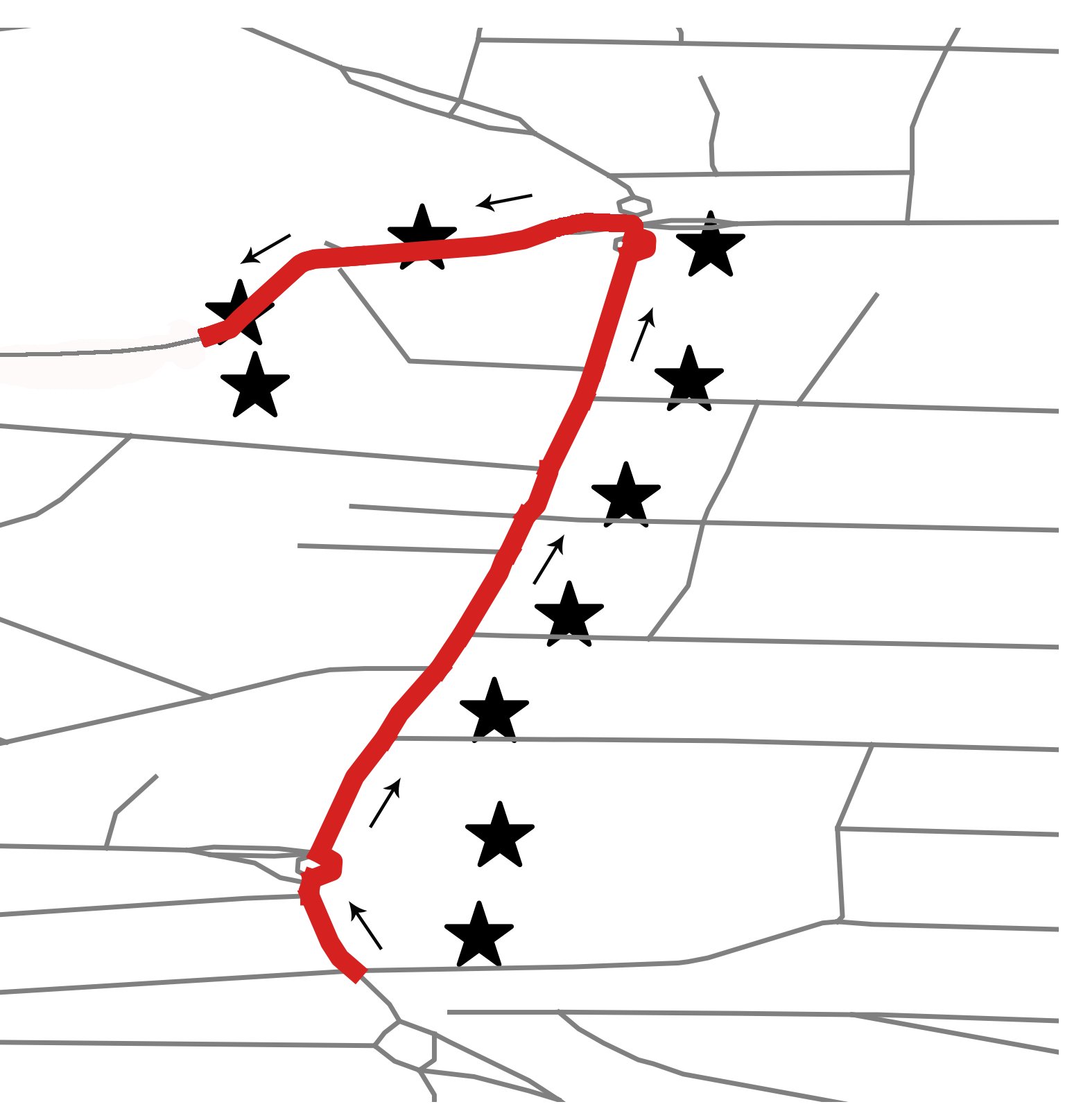}
        \caption{most likely}
        \label{probabilistic_map_matching1}
    \end{subfigure}
    \hspace{3em}
    \begin{subfigure}[b]{0.45\textwidth}
        \includegraphics[height = 6 cm,left]{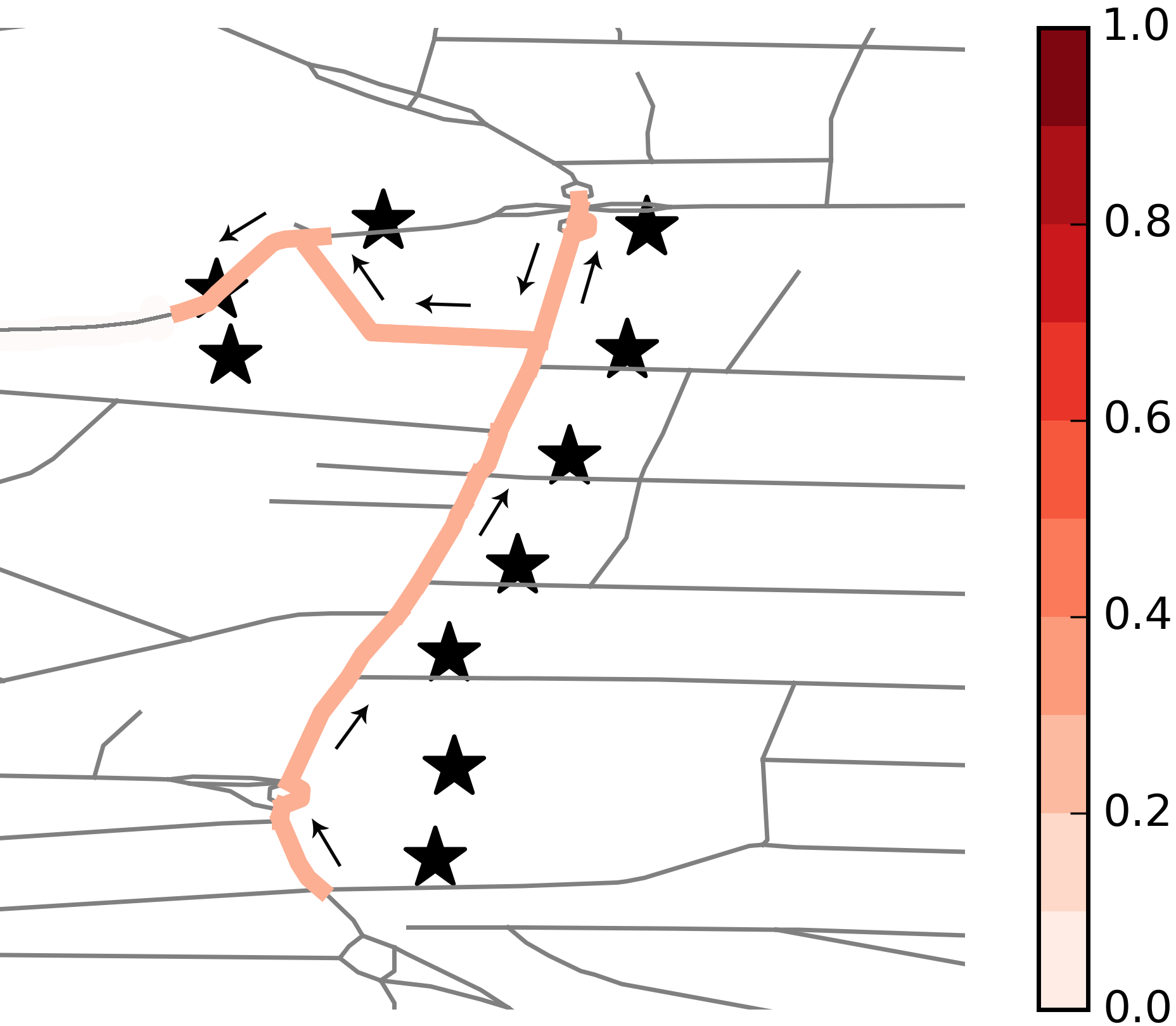}
        \caption{second most likely}
        \label{probabilistic_map_matching2}
    \end{subfigure}
    \caption{Exemplary map-matching outcome with colour-coded probability scores for the two most probable paths.}\label{probabilistic_map_matching}
\end{figure}

In the first instance, the proposed algorithm is applied to the police vehicle data. An exemplary output of the algorithm is shown in Figure~\ref{probabilistic_map_matching}. The median cross-validation error is 4.9 meters, i.e. the inferred paths tend to be 4.9 meters away from GPS points not included in the map-matching. The error approximately equals the measurement noise of the GPS data themselves, therefore the results seem to be accurate.

\begin{figure}
    \centering
    \begin{subfigure}[b]{0.45\textwidth}
        \includegraphics[width=\textwidth]{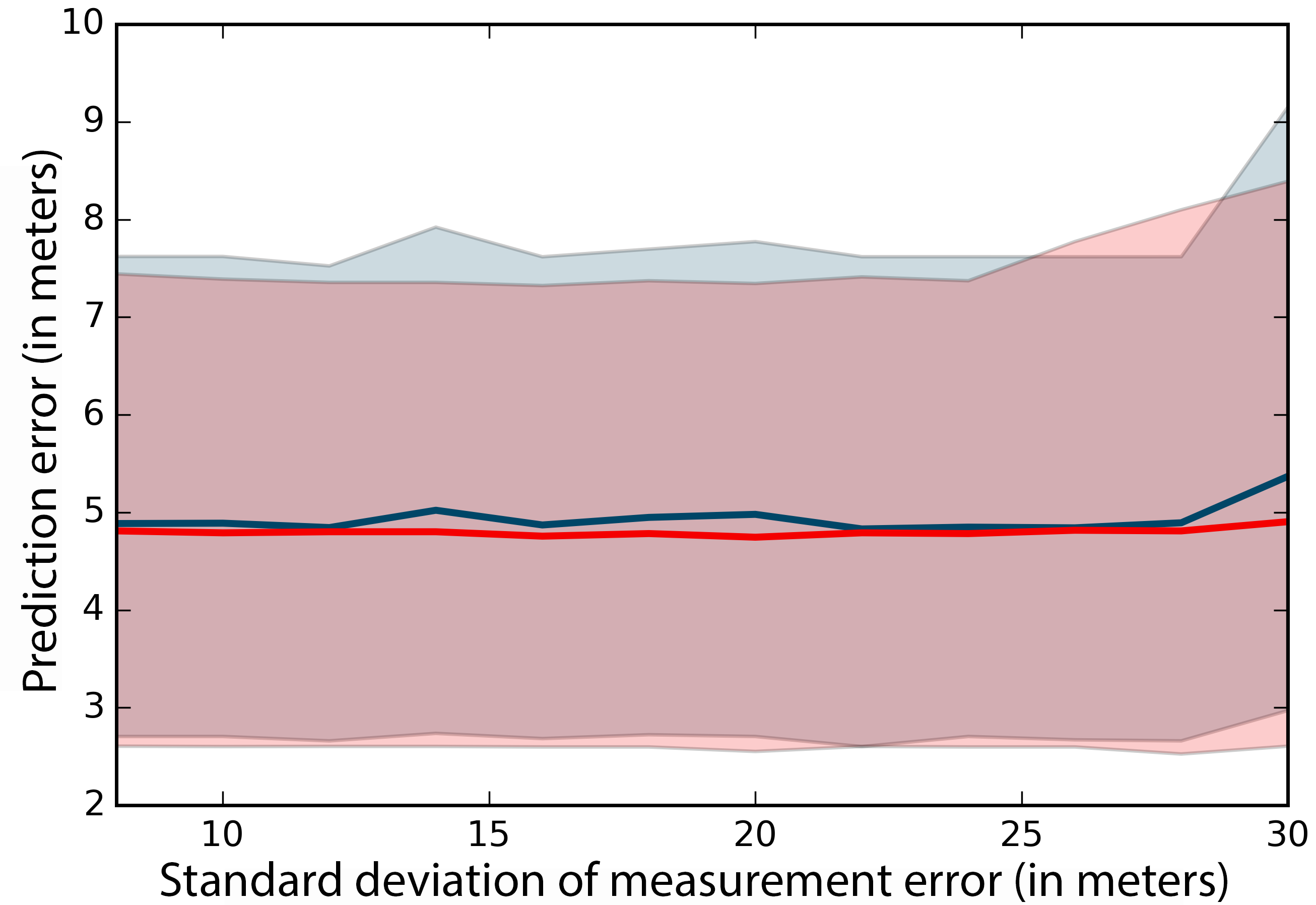}
        \caption{measurement error}
        \label{measurement_error}
    \end{subfigure}
    \hspace{1em}
    \begin{subfigure}[b]{0.45\textwidth}
        \includegraphics[width=\textwidth]{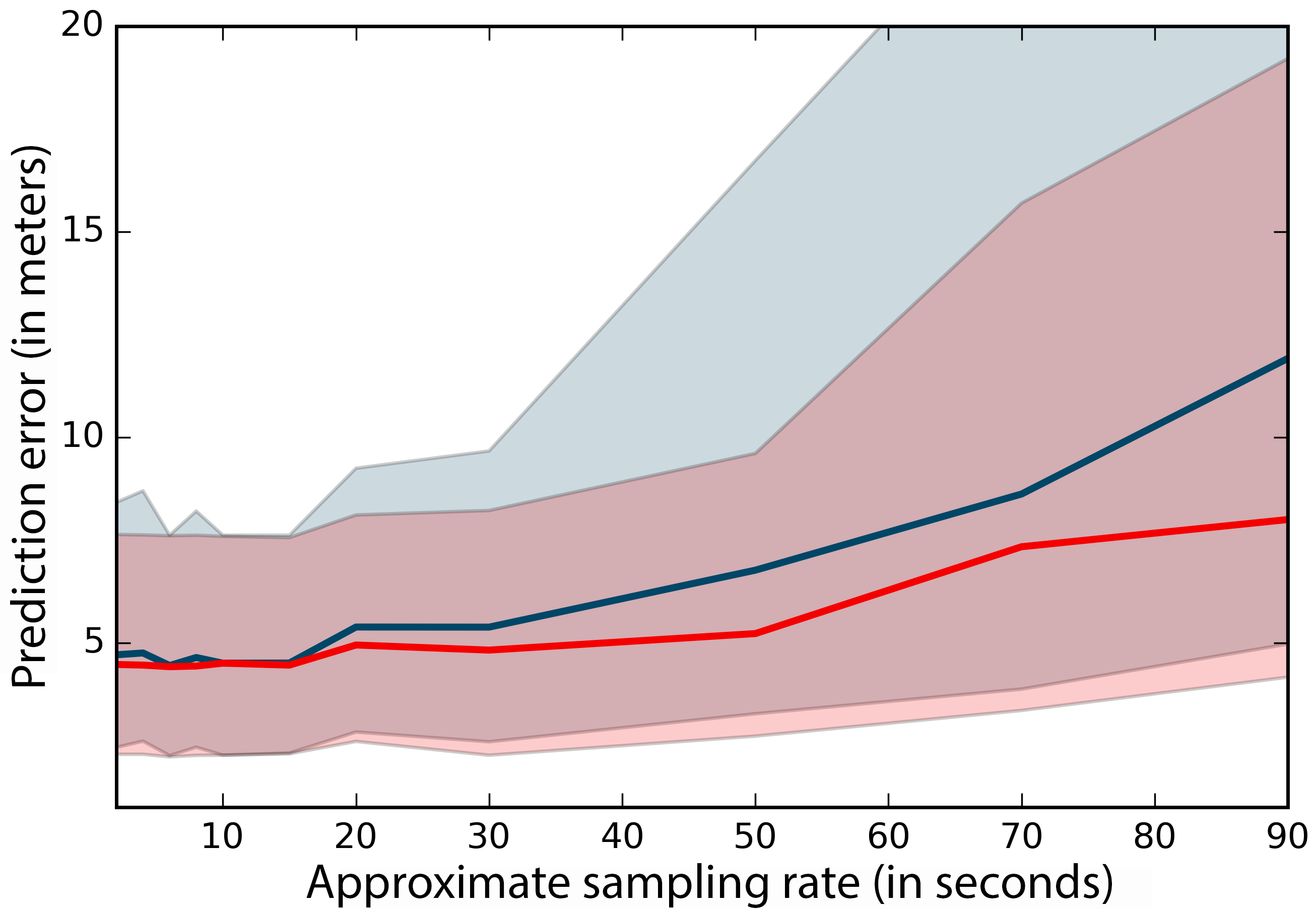}
        \caption{sampling rate}
        \label{sampling_rate}
    \end{subfigure}
    \caption{Sensitivity of Particle Filter (blue) and ST-Matching (red) to GPS measurement error and sampling rate represented as 25th, 50th and 75th percentiles of map-matching errors.}\label{pf_st_accuracy}
\end{figure}

The applicability of the algorithm to other datasets is then tested by artificially reducing the sampling rate of the data (removing some GPS points) and by increasing the noise of the data (perturbing GPS points). The algorithm shows good robustness against variation of the measurement noise (Figure~\ref{measurement_error}) that might in reality be due to high buildings, weather, etc.. However, it performs poorly on datasets with low sampling rates (Figure~\ref{sampling_rate}). The decreased performance can be explained by the fact that low sampling rates largely increase the number of possible paths that the vehicle could have taken between subsequent GPS measurements (too many to cover with a fixed number of particles). The decrease in the algorithm's performance is particularly apparent when compared to the relatively good performance of the state-of-the-art deterministic approach, the ST-Matching algorithm. 

Further work is already being undertaken to bring together strengths of the two algorithms into a highly accurate, yet fully probabilistic, map-matching algorithm. 

\section{Acknowledgements}
This work is part of the project - Crime, Policing and Citizenship (CPC): Space-Time Interactions of Dynamic Networks (www.ucl.ac.uk/cpc), supported by the UK Engineering and Physical Sciences Research Council (EP/J004197/1). The data provided by Metropolitan Police Service (London) is greatly appreciated.

\section{Biography}
Kira Kempinska is a PhD student in the Jill Dando Institute of Crime and Security Sciences at University College London. Her main research interests lie in the area of probabilistic machine learning and network analysis, particularly in application to crime and security issues.

Toby Davies is a Research Associate working on the Crime, Policing and Citizenship (CPC) project at UCL. His background is in mathematics, and his work concerns the application of mathematical techniques in the analysis and modelling of crime. His research interest include networks and the analysis of spatio-temporal patterns.

John Shawe-Taylor is a professor at University College London (UK) where he is the Head of the Department of Computer Science. His main research area is Statistical Learning Theory, but his contributions range from Neural Networks, to Machine Learning, to Graph Theory.


\bibliographystyle{apa}
\bibliography{map_matching_gisruk}

\end{document}